\documentclass[conference]{IEEEtran}
\IEEEoverridecommandlockouts
\usepackage[export]{adjustbox}
\usepackage{multirow}
\usepackage{float}
\usepackage{graphicx}
\usepackage{hyperref}
\graphicspath{ {./images/} }
\usepackage{subfig}
\usepackage[justification=centering]{caption}
\usepackage{cite}
\usepackage{amsmath,amssymb,amsfonts}
\usepackage{algorithmic}
\usepackage{graphicx}
\usepackage{textcomp}
\usepackage{xcolor}
\usepackage{alphalph}

\def\BibTeX{{\rm B\kern-.05em{\sc i\kern-.025em b}\kern-.08em
    T\kern-.1667em\lower.7ex\hbox{E}\kern-.125emX}}
\begin{document}

\title{Physical Attribute Prediction Using Deep Residual Neural Networks
}

\author{\IEEEauthorblockN{1\textsuperscript{st} Rashidedin Jahandideh}
\IEEEauthorblockA{\textit{dept. Computer Science} \\
\textit{Shahid Beheshti University}\\
Tehran, Iran \\
r.jahandideh@mail.sbu.ac.ir}
\and
\IEEEauthorblockN{2\textsuperscript{nd} Alireza Tavakoli Targhi}
\IEEEauthorblockA{\textit{dept. Computer Science} \\
\textit{Shahid Beheshti University}\\
Tehran, Iran \\
a\_tavakoli@sbu.ac.ir}
\and
\IEEEauthorblockN{3\textsuperscript{rd} Maryam Tahmasbi}
\IEEEauthorblockA{\textit{dept. Computer Science} \\
\textit{Shahid Beheshti University}\\
Tehran, Iran \\
m\_tahmasbi@sbu.ac.ir}
}

\maketitle

\begin{abstract}
Images taken from the Internet have been used alongside Deep Learning for many different tasks
such as: smile detection, ethnicity, hair style, hair colour, gender and age prediction. After
witnessing these usages, we were wondering what other attributes can be predicted from facial
images available on the Internet. In this paper we tackle the prediction of physical attributes from face images using Convolutional Neural Networks trained on our dataset named FIRW.  We crawled around 61, 000 images from the web, then use face detection to crop faces from these real world images. We choose ResNet-50 as our base network architecture. This network was pretrained for the task of face recognition by using the VGG-Face dataset, and we finetune it by using our own dataset to predict physical attributes. Separate networks are trained for the prediction of body type, ethnicity, gender, height and weight; our models achieve the following accuracies for theses tasks, respectively: 84.58\%, 87.34\%, 97.97\%, 70.51\%, 63.99\%. To validate our choice of ResNet-50 as the base architecture, we also tackle the famous CelebA dataset. Our models achieve an averagy accuracy of 91.19\% on CelebA, which is comparable to state-of-the-art approaches.
\end{abstract}


\section{Introduction}
People's physical appearance plays a big role in our day-to-day interactions: it affects our first impression of others, it has an effect on whom we are attracted to, whom we find trustworthy, etc. Physical attribute prediction also has many different uses in social media, dating, recommender systems, and surveillance. Recognizing physical attributes in people is a rather easy task for humans, because by seeing different people and observing the changes in their appearance over time, our brains have been gradually trained to perform this task. Giving computers the ability to perform this task requires that exact same training. Our aim in this paper is to train models to perform this prediction using face images and Neural Networks.

Machine learning is a broad area of study, and it offers many approaches for solving a problem.
The advantage of deep learning over traditional machine learning algorithms is that there is no need for hand crafted features, the Deep Neural Network itself will perform feature extraction and feature engineering. 
\\
Neural Networks have helped us perform some machine learning tasks much better
than we ever could, but that is not all. Convolutional Neural Networks (CNNs), a specialized version of Deep Neural Networks, have revolutionized both machine learning and computer vision. They have shown great promise in tasks such as object recognition, object detection and object
classification. 
\\
It was with the appearance of AlexNet \cite{b2} and its success in the
ImageNet competition \cite{b3} that gave rise to the popularity of Deep Neural Networks. AlexNet won the ImageNet challenge in 2012. After that Deep Neural Networks started to get wider, deeper and more complex. To investigate the effect of network depth on accuracy, VGG was released by Karen Simonyan et al. \cite{b4}. Rasmus Rothe et al. \cite{b5} finetuned VGG for apparent age estimation and won ChaLearn LAP competition in 2015. After the success of VGG, deeper network architectures such as Inception \cite{b6} and ResNet \cite{b25} were developed .
\\
In this paper, we tackle the task of physical attribute prediction from face images. We created our own dataset containing 61, 864 images, details for which is given in table \ref{t7}. We also train 40 models to predict the attributes in the famous CelebA dataset. 
\\
Physical attribute prediction from face images is a difficult task because of the non-standard ways these photographs were taken in, difference in light conditions, angle of the pictures, distance to the camera and background noise. Our models were trained to predict body type, ethnicity, gender, height and weight on our own dataset and to predict the attributes provided in CelebA. 
\\
Our contributions in this paper are:
\begin{enumerate}
  \item Exploring the capabilities of Deep Neural networks in performing complex tasks such as height and weight classification. Training and testing is performed on real world data without performing face alignment, which further complicates the problem.
  \item Using our real world data to perform multilabel ethnicity classification, which is a more challanging task compared to the binary classification performed in related works.
  \item Training a separate Residual Neural Networkfor each attribute in CelebA, which will be available for public use. 
\end{enumerate}

The rest of the paper is organized as follows: Section \ref{related works} reviews some of the related work done in recent years. Section \ref{datasets} introduces our dataset (FIRW) and CelebA. In Section \ref{experiments} we present details of FIRW, the CelebA data set, network architectures and training details. Section \ref{results} shows the results of our experiments. Finally, Section \ref{conclusion} concludes the paper.

\section{Related Works}\label{related works}
Convolutional Neural Networks (CNNs) have been used for many different tasks such as regression, classification, object detection and recognition. The introduction of ImageNet and ILSVRC provided a great opportunity for researchers to compete with each other and come up with novel nework architectures. Over recent years we have seen network architectures such as AlexNet, VGG, GoogleNet \cite{b8} and ResNet enter this competition and introduce new ideas with their innovations and achieve great results in the competition. In some cases these base networks were used for feature extraction, and classifiers such as SVM were used along  these features to perform classification. In some other cases, finetuned versions of these networks were used for either feature extraction or to directly perform classification in an end-to-end manner. Below we will review some of the recent related works. 

\subsection{}
In \cite{b5} R. Rothe et al. trained a network for apparent age estimation. They gathered their training data by scraping images from IMDB and Wikipedia websites. VGG-16 network architecture was used to perform age estimation. They showed that treating the age estimation as a classification problem yields better results than treating it as a regression task. They achieved an $\epsilon$-error of 0.264975. Dex was the winner of the ChaLearn LAP 2015 apparent age estimation challenge.

\subsection{}

L. Wen et al. \cite{b9} used ASM to detect a number of fiducial points in each face image. Facial features such as CJWR (The ratio of cheekbone width to jaw width), WHR (the ratio of the cheekbone
width to upper facial height) and ES (Average size of eyes) were automatically extracted. These features along with different regression models (SVR, LSE and GP) were used to predict BMI. The data was splitted into two sets, set 1 was used for training and set 2 for testing, and vice versa. The results were presented as mean absolute error for different age groups and also the average for each regression model. Table
\ref{t1} shows the average results (measured by mean absolute error):

\begin{table}[htbp]
\begin{center}
\begin{tabular}{|l|l|l|l|l|l|l|}
\hline
Training Set / Test Set & \multicolumn{3}{l|}{Set 1 / Set 2} & \multicolumn{3}{l|}{Set 2 / Set 1} \\ \hline
\multirow{2}{*}{Avg.}   & SVR        & LSE       & GP        & SVR        & LSE       & GP        \\ \cline{2-7} 
                        & 3.14       & 3.21      & 3.23      & 3.14       & 3.22      & 3.20      \\ \hline
\end{tabular}
\captionof{table}{Results on BMI from \cite{b9}.}\label{t1}
\end{center}
\end{table}

\subsection{}
In \cite{b10}, Deep Neural
Networks were used to infer BMI. VGG-16 and VGG-Face were used to extract features and then
a regression model was trained on those features. Table \ref{t2} shows the Pearson r
correlations on the test set for the BMI prediction task, broken down by gender. Features extracted by VGG-Face performed better than VGG-Net features.

\begin{table}[htbp]
\begin{center}
\begin{tabular}{|l|l|l|l|}
\hline
Model                  & Male & Female & Overall \\ \hline
Face-to-BMI - VGG-Net  & 0.58 & 0.36   & 0.47    \\ \hline
Face-to-BMI - VGG-Face & 0.71 & 0.57   & 0.65    \\ \hline
\end{tabular}
\captionof{table}{Results on BMI from \cite{b10}.}\label{t2}
\end{center}
\end{table}

\subsection{}
Y. Lewenberg et al. \cite{b11} used FAD (Face Attributes
Dataset) to predict 10 most sought-after traits. Table \ref{t3} shows the results of LACNN:

\begin{table}[htbp]
\begin{center}
\begin{tabular}{|l|l|}
\hline
Train              & LACNN   \\ \hline
Gender (Male)      & 98.33\% \\ \hline
Ethnicity (White)  & 83.35\% \\ \hline
Hair Colour (Dark) & 91.69\% \\ \hline
Makeup             & 92.87\% \\ \hline
Age (Young)        & 88.83\% \\ \hline
Emotions (Joy)     & 88.33\% \\ \hline
Attractive         & 78.85\% \\ \hline
Humorous           & 69.06\% \\ \hline
Chubby             & 61.38\% \\ \hline
\end{tabular}
\captionof{table}{Physical attribute prediction results from \cite{b11}.}\label{t3}
\end{center}
\end{table}

\subsection{}
Ziwei Liu et al. \cite{b12} used Deep Neural Networks for predicting face attributes such as heavy makeup, hair colour, presence of eye glasses and facial hair. Their model is comprised of a pipeline of two networks: first Lnet locates the face region in an image and then Anet is used for
feature extraction and prediction. It was shown that the performance increases by using different
pre-training strategies for LNet and ANet. They also introduced the CelebA dataset. The results
for LNets+ANet with an average accuracy of 87\% are shown in Table \ref{t6}.
\\

With the availability of the CelebA dataset and the large number of samples, other researchers
began their work on this data set. Below is a brief review of some of the work done on the CelebA
dataset. 
\\
\subsection{}
Y. Zhong et al. \cite{b13} demonstrate that using features from the middle layers of a CNN performs better than high level features of the last layers. Features from various layers (Conv2, Conv3, Conv4, Conv5, Conv6, FC1 and FC2 layers) of a modified version of FaceNet NN were used to perform
classification by training Linear SVM classifiers. The overall reported accuracy is 89.8\%. Table \ref{t6} shows the results for each attribute.
\\
\subsection{}
E. M. Hand et al. \cite{b14} took advantage of the relations between attributes. Low layers of a Deep CNN (MCNN) were shared among all attributes, while higher layers were shared between related attributes. Fourty attributes were grouped together into smaller groups based on their locations in the face or on the head. Nine attribute groups were created based on the location of atttributes. A fully connected layer named AUX is put on top of MCNN to allow the whole network to better learn the relations between outputs of MCNN and improve the accuracy for each attribute. After adding AUX on top of the network, the weights of the trained MCNN network are frozen, and only the AUX layer is
trainable. The results for MCNN-AUX are shown in \ref{t6}.
\\
\subsection{}
X. Hou et al. \cite{b15} try to improve the quality of images generated by Variational Autoencoders. They used perceptual loss instead of pixel-by-pixel loss for this improvement. To show their model’s capability in capturing semantic and conceptual information in face images, they used latent vectors from their network to perform attribute prediction on CelebA dataset. Ground truth landmark points were used to crop faces from the dataset, instead of face detectors. Latent vectors are extracted by feeding these cropped faces into the encoder network. These latent vectors are then used to train Linear SVMs to perform attribute classification. The result of their VAE-345 model with an average accuracy of 88.73\% can be seen in Table \ref{t6}.
\\
\subsection{}
R. Ranjan et al. \cite{b16} used a single network to perform face detection, face alignment, pose estimation, gender recognition, smile detection, age estimation and face recognition. A network trained for face recognition was used to initialize their CNN. To perform attribute prediction, this network branches out from different layers based
on each attribute's dependency on local or global information of the face. Several different datasets were used for training: Casia \cite{b28} (for identification and gender), MORPH \cite{b29} (for age and gender), IMDB-WIKI \cite{b5} (for age and gender), Audience \cite{b30} (for age), CelebA \cite{b12} (for gender and smile), ALFW \cite{b31} (for detection, pose and fiducials).
Their work achieves 99\% accuracy in gender classification and 93\% in smile classification on
CelebA, 93.12\% and 90.83\% on Faces of the World for gender and smile respectively.
\\
\subsection{}
Y. Zhong et al. \cite{b17} used representations extracted from levels of CNNs, which were trained for face recognition, to perform attribute prediction. Binary linear SVMs were trained for each representation to perform classification. The average accuracies were 86.6\% and 84.8\% for CelebA and LFWA respectively. The results for CelebA are shown in the Table \ref{t6}.
\\
\subsection{}
R. Torfason et al. \cite{b18} use a subset of attributes to predict
other attributes to show the correlation between attributes. 1, 10, 20, 30 and 39 attributes were used to predict the missing attribute(s). An average accuracy of 87.44\% was achieved when using 39 attributes to predict the missing attribute. Table \ref{t4} shows the results.

\begin{table}[htbp]
\begin{center}
\begin{tabular}{|l|l|l|l|l|l|l|}
\hline
\begin{tabular}[c]{@{}l@{}}Number of \\ training attributes\end{tabular} & 39                                                      & 30                                                      & 20                                                      & 10                                                       & 1                                                        & Flat Prediction                                          \\ \hline
SVM                                                                      & \begin{tabular}[c]{@{}l@{}}87.44 ± \\ 7.77\end{tabular} & \begin{tabular}[c]{@{}l@{}}86.76 ± \\ 8.03\end{tabular} & \begin{tabular}[c]{@{}l@{}}85.98 ± \\ 8.65\end{tabular} & \begin{tabular}[c]{@{}l@{}}84.23 ± \\ 10.50\end{tabular} & \begin{tabular}[c]{@{}l@{}}80.76 ± \\ 14.27\end{tabular} & \begin{tabular}[c]{@{}l@{}}80.04 ± \\ 15.35\end{tabular} \\ \hline
LASSO                                                                    & \begin{tabular}[c]{@{}l@{}}87.18 ± \\ 7.72\end{tabular} & \begin{tabular}[c]{@{}l@{}}87.11 ± \\ 8.18\end{tabular} & \begin{tabular}[c]{@{}l@{}}85.92 ± \\ 8.95\end{tabular} & \begin{tabular}[c]{@{}l@{}}84.20 ± \\ 10.32\end{tabular} & \begin{tabular}[c]{@{}l@{}}80.76 ± \\ 14.27\end{tabular} & \begin{tabular}[c]{@{}l@{}}80.04 ± \\ 15.35\end{tabular} \\ \hline
\end{tabular}
\captionof{table}{Results from \cite{b18}. Different number of attributes are used in predicting the missing attribute.}\label{t4}
\end{center}
\end{table}

They also used different combinations of handcrafted features such as LBP, SIFT, Color
Histograms and Deep Features to predict attributes. Features from the fc6 layer of a finetuned
network along with handcrafted features were used for prediction. The results for the “fc6
ft+hc+attr” combination, which achieved an average accuracy of 91\%, are reported in Table \ref{t6}.
\\
\subsection{}
D. Gao et al. \cite{b19} trained a network named ATNet\_GT to perform classifications. Motivated by the work done in \cite{b12} they also grouped the attributes together. Due the small size of their network, the 40 attributes of the CelebA dataset were grouped into 3 different groups, instead of 6. Each group was trained by branching from a different depth of the network. The final results for ATNet\_GT with an average accuracy of 90.18\% are shown in Table \ref{t6}.
\\
\subsection{}
E. M. Hand et al. \cite{2b19} propose a method called Selective Learning in order to overcome label imbalances in CelebA. Selecting Learning works on batch level during training. The goal is to have a balanced number of positive and negative samples in each batch, according to the target ditribution, by reducing the number of oversampled data and assigning more weight to undersampled data. The average accuracy for their method is 90.97\%.

\subsection{}
Y. Lu et al. \cite{2b18} use a multi-task learning framework for attribute prediction. Instead of creatng the multi-task learning network manually, they use an automatic approach. In this method, a thin version of VGG-16 network architecture, gets dynamically wider in a greedy manner during training. The authors experimented with several methods, the average accuracies of which are shown in the table below.

\begin{table}[htbp]
\begin{center}
\begin{tabular}{|l|l|}
\hline
Method              & Accuracy \\ \hline
VGG-16 Baseline     & 91.44    \\ \hline
Low-rank Baseline   & 90.88    \\ \hline
Baseline-thin-32    & 89.96    \\ \hline
Branch-32-1.0       & 90.74    \\ \hline
Branch-32.2.0       & 90.90    \\ \hline
Branch-64-1.0       & 91.26    \\ \hline
Joint Branch-32.20  & 90.4     \\ \hline
Joint Branch-64-2.0 & 91.02    \\ \hline
\end{tabular}
\captionof{table}{Results from \cite{2b18}.}\label{t4b}
\end{center}
\end{table}

\subsection{}
H. Han et al. \cite{3b19} use Deep Multi-Task Learning for attribute estimation. The proposed method learns shared features at lower levels of the network, and category-specific features at upper levels of the network. They also introduced the LFW+ dataset in their work. They achieve an average accuracy of 93\%. Detailed results are shown in table \ref{t6}.
\\

Table \ref{t5} shows some of the recent papers on the CelebA dataset. Detailed results of these papers (except for \cite{2b18} and \cite{2b19}, since results for individual attributes were not present in correspoding papers) are presented in Table \ref{t6}.

\begin{table}[htbp]
\begin{center}
\begin{tabular}{|l|l|l|}
\hline
Paper Number & Year & Title                                                                                                                             \\ \hline
1            & 2015 & \begin{tabular}[c]{@{}l@{}}Deep Learning Face Attributes in the Wild\end{tabular}                                               \\ \hline
2            & 2016 & \begin{tabular}[c]{@{}l@{}}Leveraging Mid-Level Deep\\ Representations for Predicting\\ Face Attributes in the Wild\end{tabular}  \\ \hline
3            & 2016 & \begin{tabular}[c]{@{}l@{}}Attributes for Improved Attributes:\\ A Multi-Task Network for Attribute\\ Classification\end{tabular} \\ \hline
4            & 2016 & \begin{tabular}[c]{@{}l@{}}Deep Feature Consistent Variational\\ Autoencoder\end{tabular}                                         \\ \hline
5            & 2016 & \begin{tabular}[c]{@{}l@{}}Face Attribute Prediction Using\\ Off-The-Shelf Deep Learning\\ Networks\end{tabular}                  \\ \hline
6            & 2016 & \begin{tabular}[c]{@{}l@{}}From face images and attributes\\ to attributes\end{tabular}											\\ \hline
7            & 2016 & \begin{tabular}[c]{@{}l@{}}Fully-adaptive Feature Sharing\\ in Multi-Task Networks with Applications \\ in Person Attribute Classification
\end{tabular}                                           \\ \hline
8            & 2017 & \begin{tabular}[c]{@{}l@{}}Face Attribute Prediction with\\ Convolutional Neural Networks\end{tabular}                            \\ \hline
9            & 2017 & \begin{tabular}[c]{@{}l@{}}Heterogeneous Face Attribute Estimation: \\ A Deep Multi-Task Learning Approach
\end{tabular}                            \\ \hline

10            & 2018 & \begin{tabular}[c]{@{}l@{}}Doing the Best We Can with What We Have:\\
Multi-Label Balancing with Selective Learning\\ for Attribute Prediction \end{tabular}                            \\ \hline
\end{tabular}
\captionof{table}{Recent papers on CelebA dataset.}\label{t5}
\end{center}
\end{table}

\section{Datasets}\label{datasets}
There are several datasets of face images which can be used for supervised learning, such as IMDB-Wiki \cite{b5} for gender and age prediction, Apparent Age \cite{b20} for apparent age recognition, CelebA and LFWA \cite{b12} for attribute prediction.
\\
CelebA and LFWA datasets cover a broad range of attributes. After studying the work done on
these datasets, we were wondering what other physical attributes can CNNs predict? Are CNNs
capable of predicting more complex physical attributes such as weight and height just by looking at face images?
\\
\\
Since to the best of our knowledge there were no publicly available data sets to provide height and weight attributes, we scraped the Internet and gathered our own data. Our data set, FIRW (Faces In the Real World), is not limited to height and weight, we also gathered information about body type, ethnicity and gender. Subsection \ref{ourdataset} introduces our dataset and subsection \ref{celeba} is a quick overview of CelebA.
\\
\subsection{FIRW (Faces In the Real World)}\label{ourdataset}
Our web crawler is written in Python. We used Beautiful Soup \cite{b21} for HTML parsing and Selenium Web Driver \cite{b22} to load web pages and handle web browsing. The result is a dataset containing 61, 864 images.
\\
Since some people prefered not to share some information, an image may have
all the attributes or a subset of them associated with it. For example, an image may have height and ethnicity attributes but no weight attribute. For each attribute of interest, we created a separate list to train a separate model for that attribute, so that we could benefit from all the available data. Height and weight values in the dataset are specified in range.

Table \ref{t7} gives a summary of the classes in FIRW and Table \ref{t8} shows number of samples for each attribute:

\begin{table}[htbp]
\begin{center}
\tiny
\begin{tabular}{|l|l|l|l|l|l|l|l|}
\hline
Attribute & \multicolumn{7}{l|}{Classes}                                                                                                  \\ \hline
Body type & \multicolumn{2}{l|}{Average} & \multicolumn{2}{l|}{Curvy}       & \multicolumn{3}{l|}{Large}                                  \\ \hline
Gender    & \multicolumn{3}{l|}{Female}                & \multicolumn{4}{l|}{Male}                                                        \\ \hline
Ethnicity & Black         & Asian        & White       & Indian             & Hispanic/Latino      & Middle Eastern   & Native American   \\ \hline
Height    & \multicolumn{3}{l|}{$<$ 5'7}               & \multicolumn{2}{l|}{5'7 - 6'1}            & \multicolumn{2}{l|}{$\geq$ 6'1}      \\ \hline
Weight    & \multicolumn{3}{l|}{$<$ 141 lbs.}          & \multicolumn{2}{l|}{141 lbs.  - 201 lbs.} & \multicolumn{2}{l|}{$\geq$ 201 lbs.} \\ \hline
\end{tabular}
\captionof{table}{Summary of the classes in FIRW.}\label{t7}
\end{center}
\end{table}

\begin{table}[htbp]
\begin{center}
\begin{tabular}{|l|l|}
\hline
Attributes & Number of Samples \\ \hline
Body Type  & 61, 258           \\ \hline
Ethnicity  & 50, 188           \\ \hline
Gender     & 61, 864           \\ \hline
Height     & 61, 663           \\ \hline
Weight     & 53, 932           \\ \hline
\end{tabular}
\captionof{table}{Number of samples for each attribute.}\label{t8}
\end{center}
\end{table}

Tables \ref{t9} and \ref{t10} show the available data based on gender and ethnicity respictively:

\begin{table}[htbp]
\begin{center}
\begin{tabular}{|l|l|l|}
\hline
Attributes & Female  & Male    \\ \hline
Body Type  & 44, 349 & 16, 909 \\ \hline
Ethnicity  & 36, 550 & 13, 638 \\ \hline
Gender     & 44, 790 & 17, 085 \\ \hline
Height     & 44, 701 & 16, 962 \\ \hline
Weight     & 37, 450 & 16, 482 \\ \hline
\end{tabular}
\captionof{table}{Number of samples based on gender.}\label{t9}
\end{center}
\end{table}

\begin{table}[htbp]
\begin{center}
\tiny
\begin{tabular}{|l|l|l|l|l|l|l|l|}
\hline
Attributes & Black  & Asian  & White   & Indian & Hispanic/Latino & Middle Eastern & Native American \\ \hline
Body Type  & 2, 445 & 1, 801 & 41, 364 & 728    & 2, 720          & 382            & 405             \\ \hline
Ethnicity  & 2, 459 & 1, 823 & 41, 636 & 729    & 2, 749          & 385            & 407             \\ \hline
Gender     & 2, 459 & 1, 823 & 41, 636 & 729    & 2, 749          & 335            & 407             \\ \hline
Height     & 2, 355 & 1, 819 & 41, 594 & 724    & 2, 745          & 383            & 407             \\ \hline
Weight     & 2, 169 & 1, 711 & 36, 111 & 683    & 2, 511          & 356            & 355             \\ \hline
\end{tabular}
\captionof{table}{Number of samples based in ethnicity.}\label{t10}
\end{center}
\end{table}

Fig. \ref{f1} shows some samples from FIRW:

\begin{figure}[htbp]
\centerline{\includegraphics[width=8.5cm,height=8.5cm,keepaspectratio]{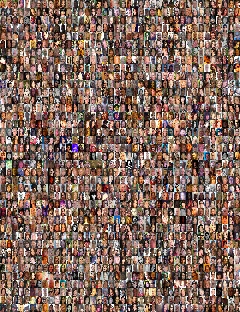}}
\caption{Sample images from the dataset. (High quality images are not provided to protect user privacy).}
\label{f1}
\end{figure}

\subsection{CelebA}\label{celeba}
\begin{figure}[htbp]
\centerline{\includegraphics[width=10cm,height=10cm,keepaspectratio]{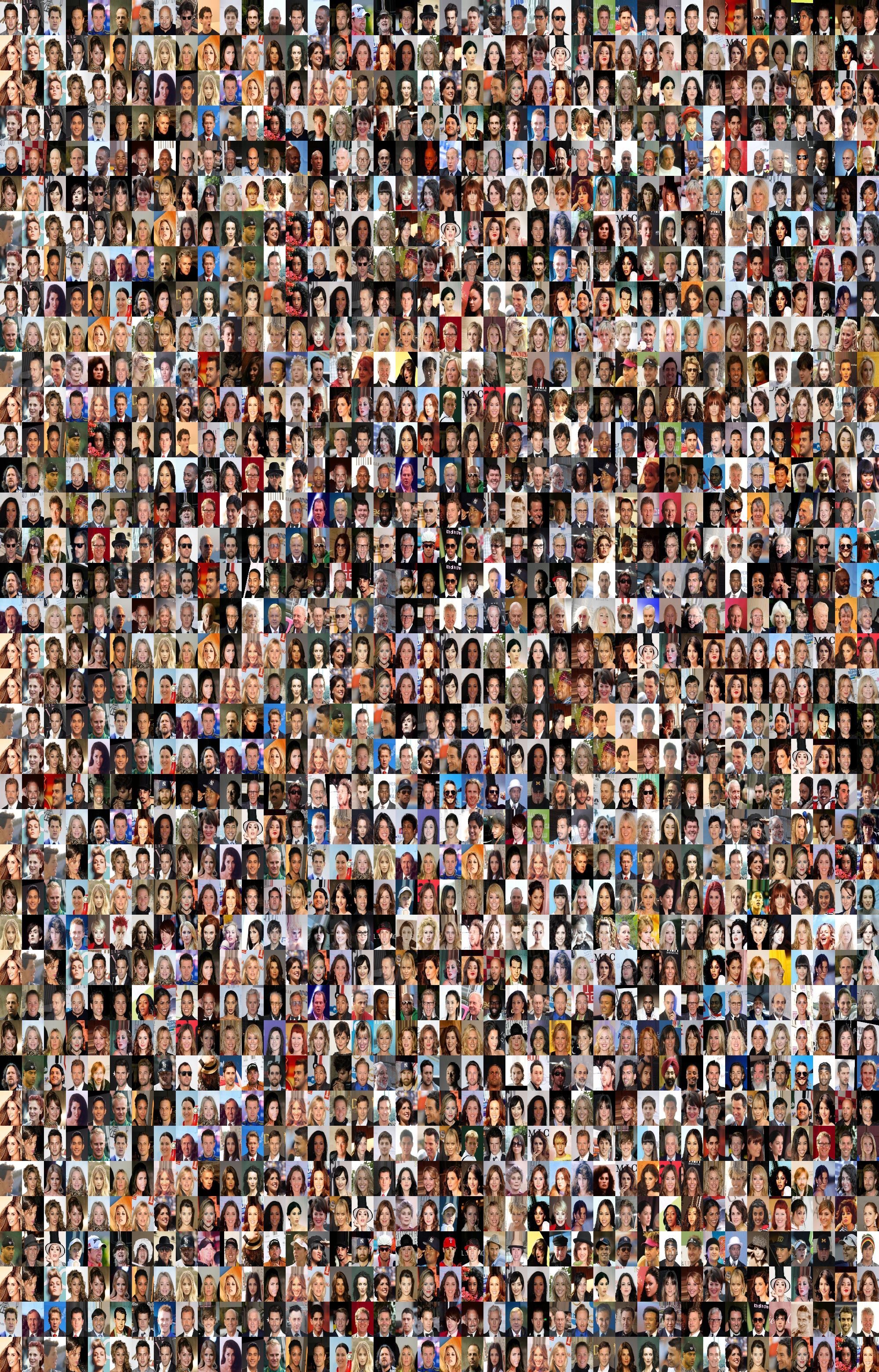}}
\caption{Samples images from CelebA dataset. (Each row represents true samples for each attribte).}
\label{f2}
\end{figure}

Z. Liu et al. \cite{b12} introduced CelebA dataset. CelebA consists of 202, 599 images. Each image is annotated with 40 attributes. Table \ref{t6} shows the attributes in CelebA and Fig.\ref{f2} shows samples from CelebA dataset.

\section{Experiments}\label{experiments}
To see the performance of Residual Neworks in predicting attributes, we used ResNet-50 to perform our experiments on two datasets: 
\begin{enumerate}
  \item FIRW, introduced in \ref{ourdataset}.
  \item CelebA dataset, introduced in \ref{celeba}.
  \\
\end{enumerate}
First, we will briefly review the motivation behind Residual Networks in \ref{DeepResidualNeuralNetworks}, and subsection \ref{OE} presents our experiments.

\subsection{Deep Residual Neural Networks}\label{DeepResidualNeuralNetworks}
It was the success of AlexNet in using Deep Neural Networks to perform classification that caught the attention of reserchers. Over time, the focus shifted from smaller and shallower architectures such as AlexNet, to deeper networks like VGG. 
To see the effect of network depth on performance and accuracy, K. Simonyan et al. \cite{b24} started stacking convolutional layers on top of each other, while other network parameters were kept fixed. This resulted in various VGG models with different number of layers, from 11 to 19. VGG models also differ from their previous counterparts in the size of receptive fields (convolutional mask) and the stride used in the network (stride is set to 1). While previous networks used larger receptive fields, such as $11 \times 11$ or $7 \times 7$, VGG models use $3 \times 3$ receptive fields throughout the whole network. Stacking several of these receptive fields on top of one another achieves the same result as using larger receptive fields (depending on the number of layers stacked together). These changes result in a stack of convolutional layers with a smaller receptive field, to have fewer parameters than their equivalent convolutional layer with a larger receptive field. The final verdict was that the increase in depth helps in achieving higher accuracy.
\\
Unfortunately because of the vanishing/exploding gradient problem, simply stacking layers on top of each other to gain better performance will not be an easy task\cite{b27}. Vanishing/exploding gradients can prevent networks from converging. Several solutions such as normalized initialization \cite{b27} were proposed to overcome this obstacle.
Even by utilizing these solutions, we encounter another problem, which is the degradation problem\cite{b25}. The degradation problem states that with the increase in network depth, accuracy might get saturated or even decrease\cite{b25}. K. He et al. \cite{b25} introduce a Deep Residula Learning framework to solve this issue. In Residual Learning, instead of learning a direct maping between inputs and outputs, the network learns a residual mapping. This goal can be achieved by using shortcut connections, which are connections that skip some layers. Shortcut connections proposed in \cite{b25}, perform identity mapping. Identity shorcut connections add no extra parameters to the model. The addition of these shorcut connections allows Deep Residual Networks to achieve higher accuracy compared to networks with the exact architecture of Residual Networks except for the omission of shortcut connections\cite{b25}.
\\
K. He et al. \cite{b25} entered the ImageNet competition with Residual Networks with depth of 18, 34, 50, 102 and 152. An ensemble of six Deep Residual Networks won the 1st place in ILSVRC 2015.
 
\subsection{Our Experiments}\label{OE}
When a deep neural network is trained from the scratch, the weights in its layers are set randomly
in the beginning and are updated during the training. Since this is a time consuming task and may
cause overfitting when the size of the training set is not large enough, we use pre-trained networks
and fine tuning to train our networks. Pre-trained networks are networks that have been trained on
large datasets for classification and recognition tasks. These pre-trained networks and their weights can be used as a base network to perform other prediction and classification tasks and can be finetuned using the dataset associated with the task in hand. By using them, we will not be starting with random weight initialization and use their weights instead.  
We use networks and weights provided by the VGG-Face implementation in Keras \cite{b26}. This network was trained for face recognition on the VGG-Face dataset.
\\

Subsections \ref{ourexperiments} and \ref{celebaexperiment} present the experiment details on FIRW and CelebA, respectively. 
\\
\subsubsection{FIRW}\label{ourexperiments}
Predicting features based on profile picture is a daunting task because of the different conditions profile pictures were taken in, bad lighting conditions, occlusions, distance to the camera, surrounding environment and other factors. After data gathering and preparation, we performed face detection on images and cropped the face from the image. We used an off-the-shelf face recognition and manipulation library based on dlib and Deep Learning for face detection \cite{b23}. When a face was detected in an image, we added \%40 padding to left, right and upper borders and \%30 to the bottom part of the detected face whenever possible. If these paddings were not possible, we tried padding with smaller values for each border of the face bounding box. These cropped faces are then saved to be used during training, validation and testing. These cropped face images may have diferrent sizes, so all images are resized before being fed to the network, by specifying a target size of $224 \times 224$ in Keras \cite{2b23} ImageDataGenerators, for training and validation. We use data augmentation during the training phase. During validation and test phases, images are only resized and rescaled before going through the models for prediction. 
\\
We want to train our models on real world data, so we did not perform any face alignment on the images before feeding them to our models. This results in a more challenging classification task compared to the CelebA dataset. We also believe that since images on CelebA are from celebrities, they are taken in better conditions compared to images in FIRW, which further complicates the classification task.
\\
For each label, a separate ResNet-50 \cite{b25} network was trained. The top parts of these networks are replaced by a Flatten layer and a Dense Layer with several Softmax nodes, equal to the number of classes in the task at hand, for example 3 nodes for height classification and 7 for ethnicity classification. We finetune these networks using our dataset. 

Before starting the training process, \%20 of the data for each label was set aside to be used for
testing. For validation, \%20 of training data was used.
For each attribute a separate model was trained. 
\\
We chose “categorical crossentropy” as the loss
function and “accuracy” as the metric to train the networks on our own dataset. Our choice of
optimizer is the “Adam optimizer” with the default configuration. 
\\
Each network was trained for 100 epochs with a batch size of 32. Keras callbacks were used to save the model with the best performance on the validation data, and then these saved models were used to perform predictions on the test data. 
\\

\subsubsection{CelebA}\label{celebaexperiment}
As mentioned in \ref{celeba}, there are 202,599 images in CelebA dataset each annotated with 40 attributes. Train, validation and test sets were created based on the specified values in CelebA webpage. Aligned images available in the webpage are used for training, validation and testing. Data augmentaion is used during the training to prevent the networks from overfitting. For validation and test phases, images are resized and rescaled. Input size for all networks is $224 \times 224$.

For each attribute, a separate model was trained. Once again we used ResNet-50 architecture provided in the Keras VGG-Face library, and finetuned it using CelebA dataset. The top parts of the networks are replaced by a Flatten layer and a Dense layer with a single Sigmoid node. Training configuration is as follows: “binary crossentrpy” as loss function, “Adam optimizer” with the default configuration as the optimizer and “accuracy” as the metric. Due to the bigger size of CelebA dataset we trained each model for 50 epochs with a batch size of 32. Keras callbacks were used to save the model with the best performance on the validation data, and these model were used to perform prediction on the test data.

All the experiments, both on FIRW and CelebA, were performed on a system running Ubuntu 16.04 with an nVidia GTX 1080Ti GPU.

\section{Results}\label{results}
\subsection{FIRW}
We have trained separate ResNet-50 models for each attribute in our dataset. Table \ref{t12} shows the results of these models:

\begin{table}[htbp]
\begin{center}

\begin{tabular}{|l|l|}
\hline
Attributes & Accuracy (\%) \\ \hline
Body Type  & 84.58         \\ \hline
Ethnicity  & 87.34         \\ \hline
Gender     & 97.97         \\ \hline
Height     & 70.51         \\ \hline
Weight     & 63.99         \\ \hline
\end{tabular}
\captionof{table}{Results on FIRW.}\label{t12}
\end{center}
\end{table}

\begin{figure}[htbp]
  \centering
  \subfloat[Body Type]{\label{ref_label2}\includegraphics[scale=0.3]{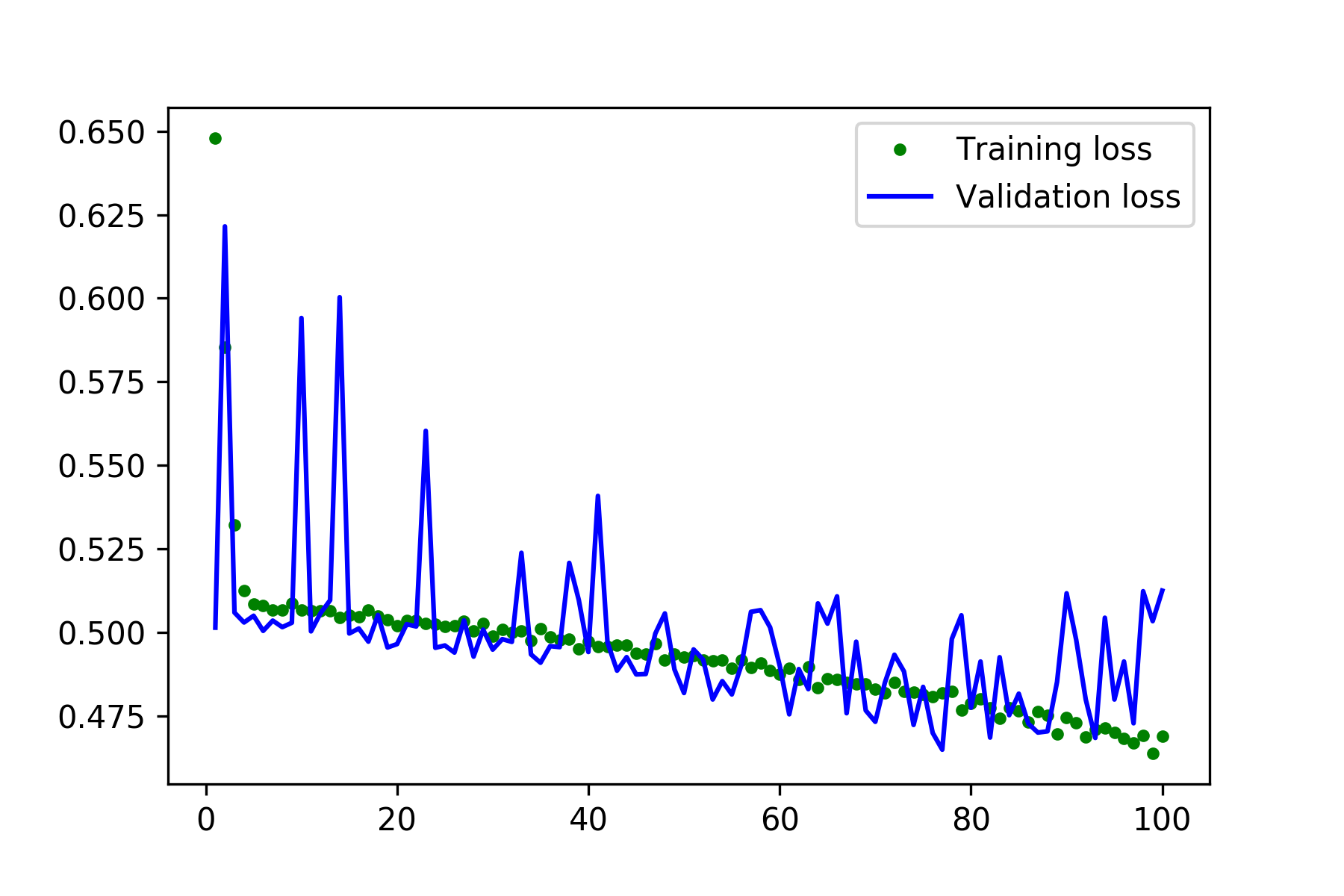}}
  \subfloat[Ethnicity]{\label{ref_label1}\includegraphics[scale=0.3]{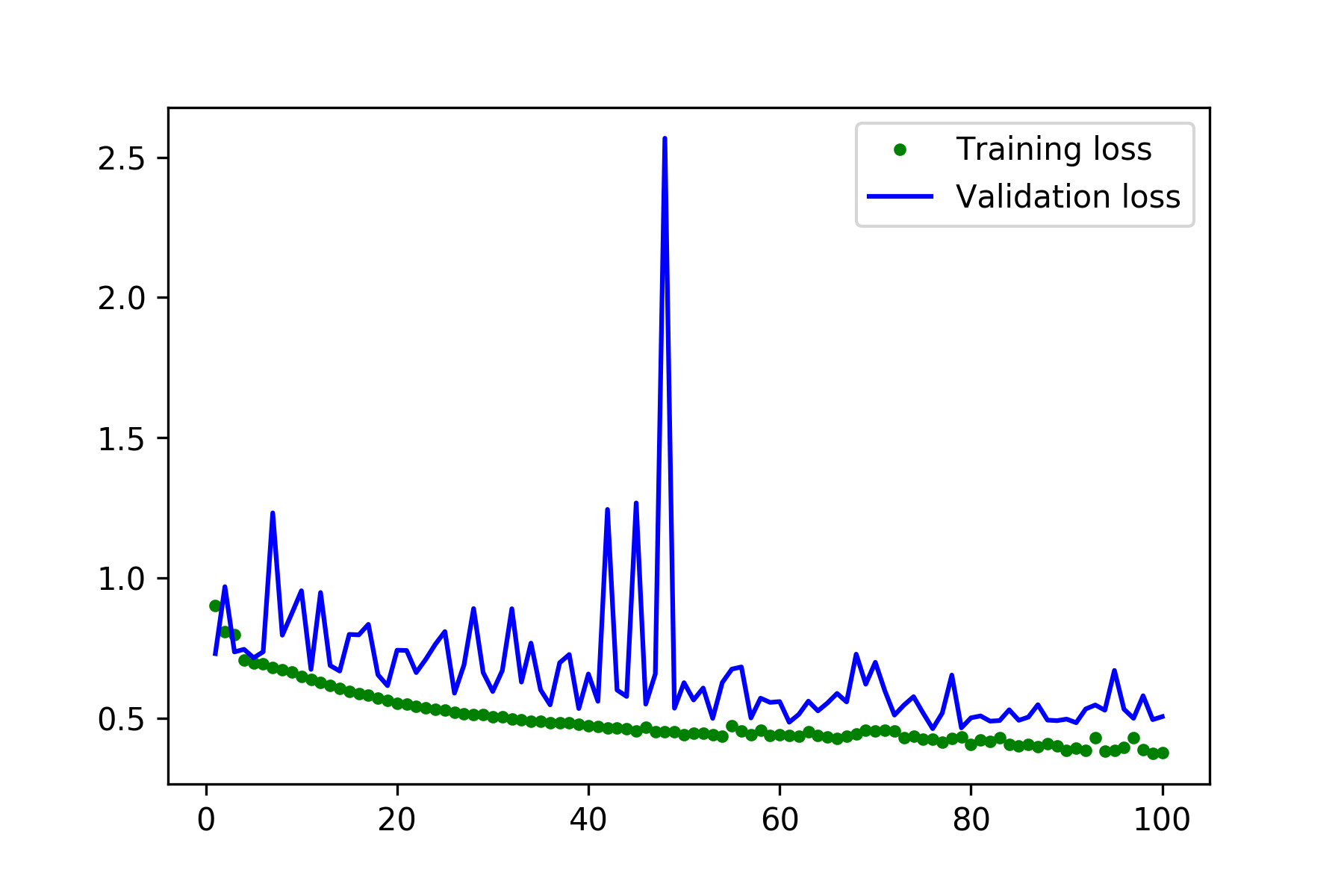}}
  \\ 
  \subfloat[Gender]{\label{ref_label2}\includegraphics[scale=0.3]{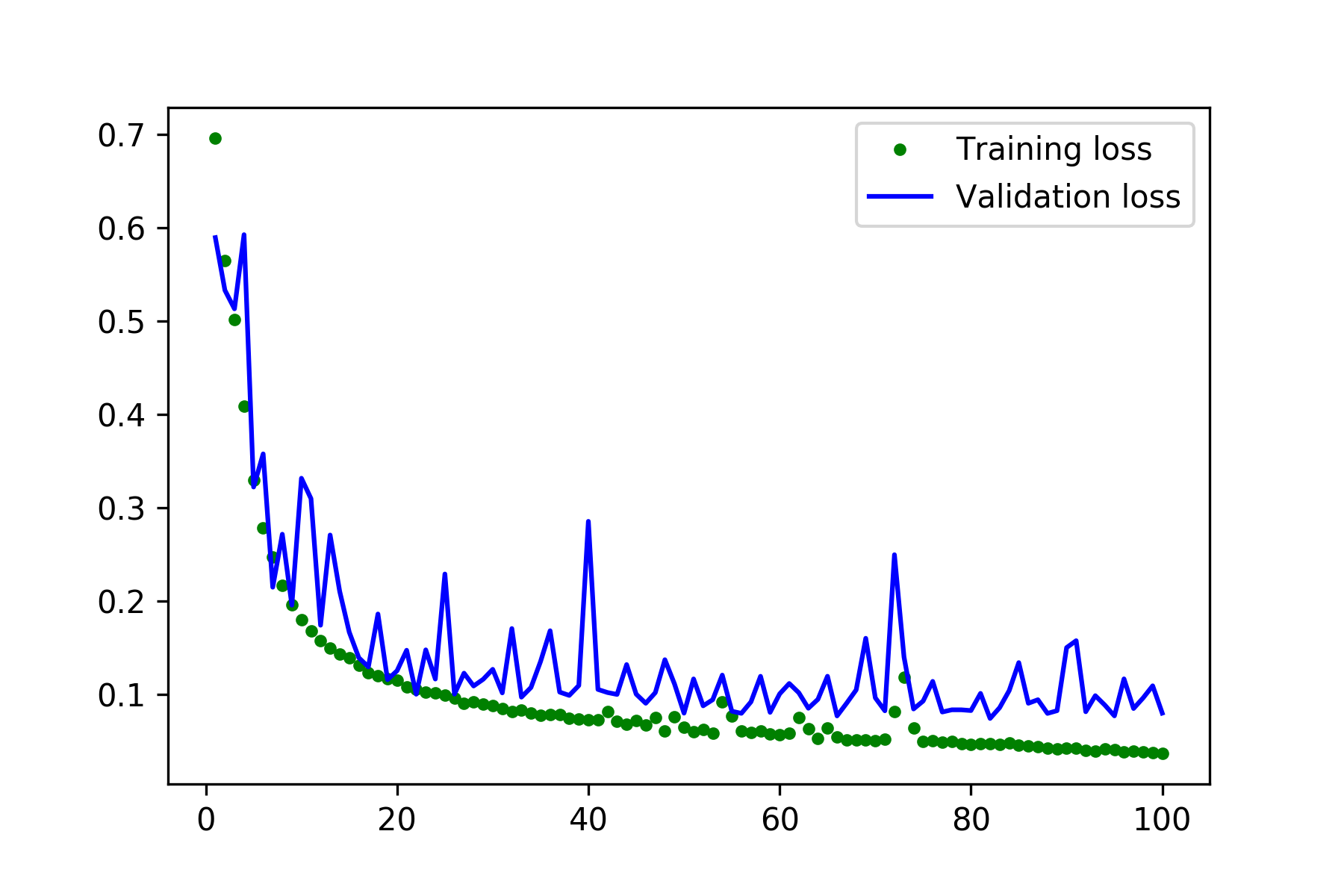}} 
  \subfloat[Height]{\label{ref_label1}\includegraphics[scale=0.3]{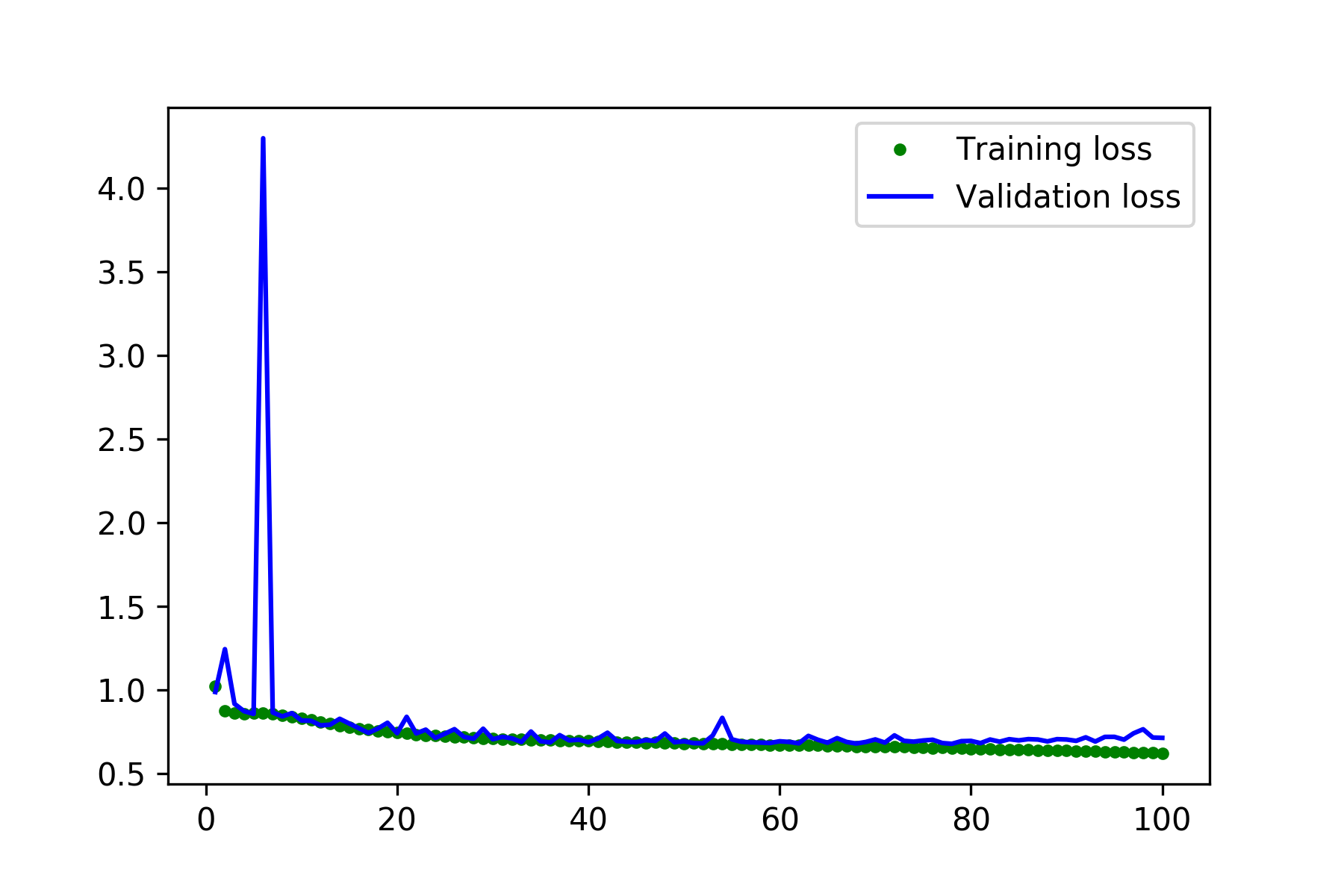}}
  \\ 
  \subfloat[Weight]{\label{ref_label2}\includegraphics[scale=0.3]{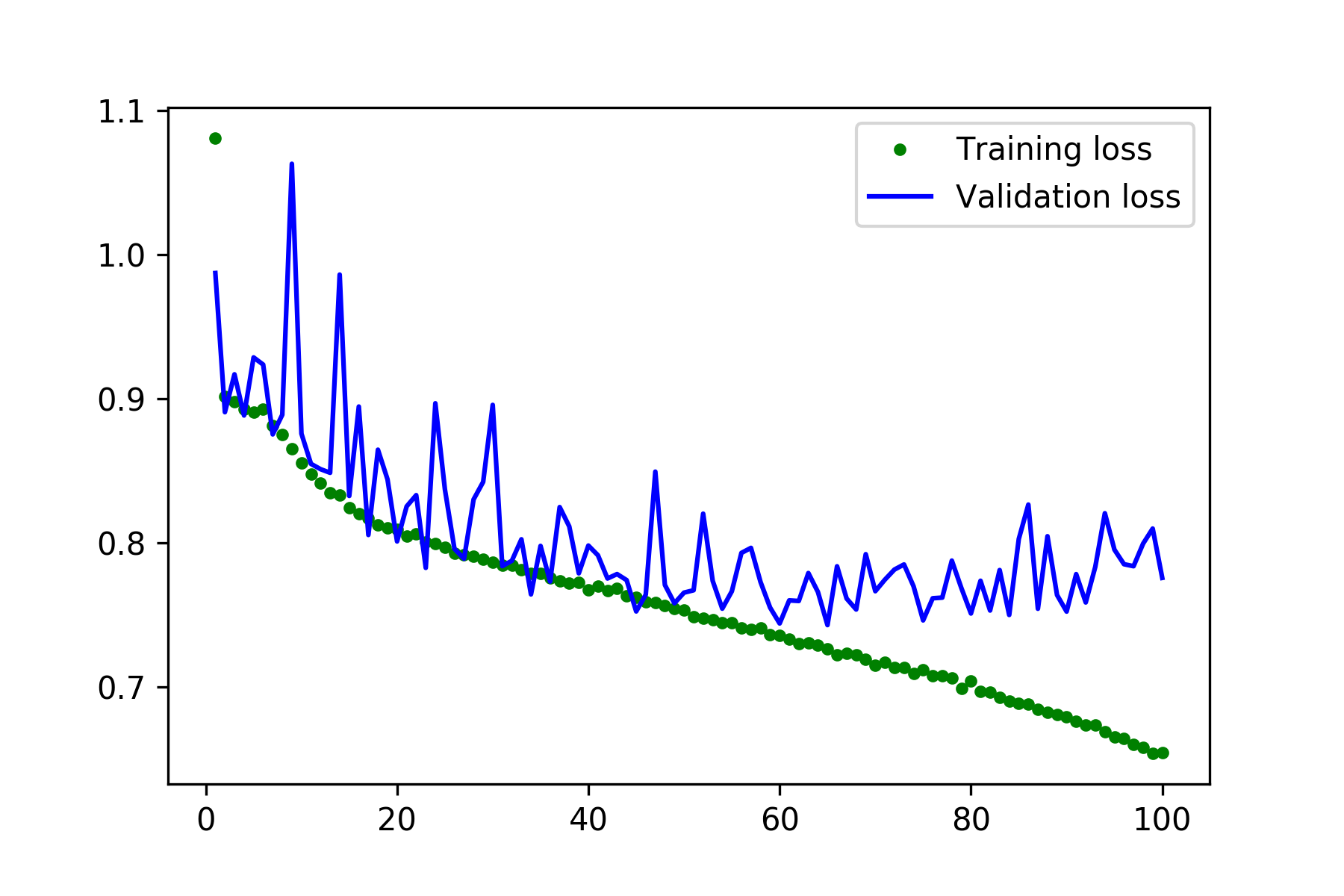}}
  \caption{\label{ref_label_overall}Loss Histories of ResNet-50 models trained on FIRW.}
  \label{f4}
\end{figure}

Fig. \ref{f4} shows the training history of each ResNet-50 model based on loss per epoch.

Except for the gender classification problem, which is a binary classification problem, all other tasks are multiclass classifications. Gender classification is popular among researchers and we have reviewed several related works devoted to this task. Even though our dataset is smaller than the datasets used in related works, ResNet-50 still manages to achieve 97.97\% accuracy.
\\

We performed a multiclass prediction in body type classification and our trained models achieve an accuracy of 84.58\%.
Considering that our model is performing multiclass classification and with fewer training samples, our model performs well in this classification task. The availablity of more training data can further improve the result.
\\ 

Height and weigh prediction from face images are challenging problems, especially considering that the labels for each image in FIRW were provided by users and that we did not perform any face alignment on images. Regarding these tasks, our ResNet-50 models achieve good results, 70.51\% and 63.99\% in height and weight prediction, respectively. We believe that with the availability of more training data with more accurate labels for height and weight, it would be possible to achieve better results.
\\
In ethnicity prediction, ResNet-50 achieves an accuracy of 86.12\%. Ethnicity prediction is a multiclass problem in our dataset and we achieve a higher accuracy than the binary classification performed in \cite{b11}.

\subsection{CelebA}
For each attribute in CelebA, a separate ResNet-50 model was trained. Table \ref{t6} shows the results we achieved for each attribute. The average accuracy of our models is 91.19\%.
\\

Our models perform classification in an end-to-end manner. Final representations learned by each ResNet-50 model are used by the network to predict the target. Our simple end-to-end networks only use the aligned \& cropped images provided on the CelebA web page. Except for data augmentation (used only on training set during training), resizing and rescaling, we did not perform any other preprocessing on images. 
\\

Table \ref{t6} shows our resluts and that of related works.

\section{Conclusion}\label{conclusion}
Our purpose in this work was to investigate what physical attributes Deep Neural Networks can predict, other than those investigated by the research community in datasets such as CelebA and LFWA.  Predicting height and weight seem to be more challenging problems. Our results showed tha ResNet-50 performs well in classifying these attributes. We achieved higher accuracies for predicting gender, body type and ethnicity compared to the more demanding tasks of height and weight classification. The availablity of more training data along with having more accurate labels for height and weight is crucial in further exploring this task. 
\\
We also used ResNet-50 in classifying CelebA attributes. Our models achieve good accuracies and are comparable to state-of-the-art approaches. 
\\
We plan to train a multi-task learning version of ResNet-50 in future to see how its performance compares with the works we have reviewed here and investigate how big of a role network depth can play in a multi-task framework.

\begin{table}[htbp]
\tiny
\begin{tabular}{|l|l|l|l|l|l|l|l|l|l|}
\hline
Paper Number          & 1  & 2     & 3     & 4  & 5  & 6     & 8 & 9 & Ours \\ \hline
5 o'Clock Shadow      & 91 & 93.34 & 94.51 & 89 & 77 & 94.87 & 92 & \textbf{95} & 94.60  \\ \hline
Arched Eyebrows       & 79 & 82.50 & 83.42 & 80 & 83 & 84.08 & 81 & \textbf{86} & 83.56  \\ \hline
Attractive            & 81 & 80.77 & 83.06 & 78 & 79 & \textbf{86.62} & 81 & 85 & 82.32  \\ \hline
Bags\_Under\_Eyes     & 79 & 82.24 & 84.92 & 82 & 83 & 85.79 & 84 & \textbf{99} & 84.95  \\ \hline
Bald                  & 98 & 97.75 & 98.90 & 98 & 91 & 98.78 & \textbf{99} & \textbf{99} & 98.95  \\ \hline
Bangs                 & 95 & 95.58 & \textbf{96.05} & 95 & 91 & 95.14 & 96 & 96 & 95.89  \\ \hline
Big\_Lips             & 68 & 69.90 & 71.47 & 77 & 78 & 73.99 & 71 & \textbf{88} & 70.21  \\ \hline
Big\_Nose             & 78 & 82.64 & 84.53 & 81 & 83 & 85.35 & 83 & \textbf{92} & 83.61  \\ \hline
Black\_Hair           & 88 & 86.04 & 89.78 & 85 & \textbf{91} & 88.60 & 89 & 85 & 90.12  \\ \hline
Blond\_Hair           & 95 & 94.89 & 96.01 & 93 & \textbf{97} & 95.77 & 95 & 91 & 96.02  \\ \hline
Blurry                & 84 & 96.15 & 96.17 & 95 & 88 & \textbf{96.24} & 96 & 96 & 95.99  \\ \hline
Brown\_Hair           & 80 & 84.15 & 89.15 & 80 & 76 & 88.04 & 87 & \textbf{96} & 88.47  \\ \hline
Bushy\_Eyebrows       & 90 & 91.89 & \textbf{92.84} & 88 & 83 & 90.75 & 92 & 85 & 92.42  \\ \hline
Chubby                & 91 & 94.87 & 95.67 & 94 & 75 & 96.16 & 94 & \textbf{97} & 95.68  \\ \hline
Double\_Chin          & 92 & 96.19 & 96.32 & 96 & 80 & 96.80 & 96 & \textbf{99} & 96.23  \\ \hline
Eyeglasses            & 99 & 99.48 & 99.63 & 99 & 91 & 98.89 & 99 & 99 & \textbf{99.67}  \\ \hline
Goatee                & 95 & 97.07 & 97.24 & 95 & 83 & 97.15 & 97 & \textbf{98} & 97.55  \\ \hline
Gray\_Hair            & 97 & 97.77 & 98.20 & 97 & 87 & 98.29 & 98 & 96 & \textbf{98.34}  \\ \hline
Heavy\_Makeup         & 90 & 90.14 & 91.55 & 89 & \textbf{95} & 91.78 & 90 & 92 & 91.29  \\ \hline
High\_Cheekbones      & 87 & 86.06 & 87.58 & 85 & \textbf{88} & 87.41 & 86 & \textbf{88} & 87.80  \\ \hline
Male                  & 98 & 98.09 & 98.17 & 95 & 94 & 97.74 & 97 & 98 & \textbf{98.24}  \\ \hline
Mouth\_Slightly\_Open & 92 & 92.56 & 93.74 & 88 & 81 & 89.27 & 93 & \textbf{94} & 93.78  \\ \hline
Mustache              & 95 & 96.56 & 96.88 & 96 & 94 & \textbf{97.25} & 97 & 97 & 96.81  \\ \hline
Narrow\_Eyes          & 81 & 86.92 & 87.23 & 89 & 81 & 85.99 & 86 & \textbf{90} & 87.32  \\ \hline
No\_Beard             & 95 & 95.38 & 96.05 & 91 & 80 & 96.38 & 94 & \textbf{97} & 96.05  \\ \hline
Oval\_Face            & 66 & 70.63 & 75.84 & 74 & 75 & \textbf{78.33} & 76 & 78 & 74.95  \\ \hline
Pale\_Skin            & 91 & 96.69 & \textbf{97.05} & 96 & 73 & 96.81 & 97 & 97 & 97.04  \\ \hline
Pointy\_Nose          & 72 & 76.17 & 77.47 & 74 & \textbf{83} & 75.60 & 75 & 78 & 77.40  \\ \hline
Receding\_Hairline    & 89 & 92.14 & 93.81 & 92 & 86 & 92.67 & 93 & \textbf{94} & 93.60  \\ \hline
Rosy\_Cheeks          & 90 & 94.29 & 95.16 & 94 & 82 & 94.82 & 95 & \textbf{96} & 95.13  \\ \hline
Sideburns             & 96 & 97.44 & 97.85 & 96 & 82 & 97.58 & 97 & \textbf{98} & 97.96  \\ \hline
Smiling               & 92 & 92.11 & 92.73 & 91 & 90 & 92.65 & 92 & \textbf{94} & 93.19  \\ \hline
Straight\_Hair        & 73 & 80.00 & 83.58 & 80 & 77 & 83.21 & 80 & \textbf{85} & 83.91  \\ \hline
Wavy\_Hair            & 80 & 77.35 & 83.91 & 79 & 77 & 84.17 & 82 & \textbf{87} & 83.24  \\ \hline
Wearing\_Earrings     & 82 & 86.74 & 90.43 & 84 & \textbf{95} & 87.25 & 89 & 91 & 90.47  \\ \hline
Wearing\_Hat          & 99 & 98.78 & 99.05 & 98 & 90 & 98.89 & 99 & 99 & \textbf{99.09}  \\ \hline
Wearing\_Lipstick     & 93 & 92.35 & 94.11 & 91 & \textbf{95} & 94.13 & 93 & 93 & 93.25  \\ \hline
Wearing\_Necklace     & 71 & 85.78 & 86.63 & 88 & \textbf{90} & 86.80 & 86 & 89 & 87.63  \\ \hline
Wearing\_Necktie      & 93 & 94.42 & 96.51 & 93 & 81 & 95.82 & 96 & \textbf{97} & 96.96  \\ \hline
Young                 & 87 & 87.48 & 88.48 & 84 & 86 & 88.92 & 88 & \textbf{90} & 88.04  \\ \hline
\textbf{Average}           	  & 87 & 89.8 & 91.29 & 88.73 & 86.6 & 91.00 & 90.18 & \textbf{93} & 91.19 \\ \hline
\end{tabular}
\captionof{table}{Detailed results of papers in table \ref{t5} (except for \cite{2b18} and \cite{2b19}) and our results on CelebA dataset.}\label{t6}
\end{table}


\begin{thebibliography}{00}
\bibitem{b1} A. H. Marblestone, G. Wayne, and K. P. Kording, ``Towards an integration of deep learning and neuroscience,'' arxiv, June 2016.

\bibitem{b2} A. Krizhevsky, I. Sutskever, G. E. Hinton, ``ImageNet Classification with Deep Convolutional Neural Networks,'' ACM Digital Library, 2012.

\bibitem{b3} J. Deng, W. Dong, R. Socher , L. Li, K. Li, and F. Li. ``ImageNet: A Large-Scale Hierarchical Image Database,'' CVPR, 2009.

\bibitem{b4} K. Simonyan, and A. Zisserman, ``Very deep convolutional networks
for large-scale image recognition,'' ICLR, 2015.


\bibitem{b5} R. Rothe, R. Timofte, and L. Van Gool, ``DEX: Deep EXpectation of apparent age from a single image,'' ICCV, 2015.

\bibitem{b6} S. Ioffe, and C. Szegedy, ``Batch normalization: accelerating deep network training by reducing internal covariate shift,'' CVPR, 2015.


\bibitem{b8} C. Szegedy, W. Liu, Y. Jia, P. Sermanet, S. Reed, D. Anguelov, D. Erhan, V. Vanhoucke, and A. Rabinovich, ``Going Deeper With Convolutions,'' CVPR, 2015.

\bibitem{b9} L. Wen, and G. Guo, ``A computation approach to body mass index prediction from face images,'' Image and Vision Computing, May 2013.

\bibitem{b10} E. Kocabey, M. Camurcu, F. Ofli, Y. Aytar, J. Marin, A. Torralba, I. Weber, ``
Face-to-BMI: Using Computer Vision to Infer Body Mass Index on Social Media,'' ICWSM, 2017.

\bibitem{b11} Y. Lewenberg, Y. Bachrach, S. Shankar, and A. Criminisi, ``Predicting Personal Traits from Facial Images Using Convolutional Neural Networks Augmented with Facial Landmark Information,'' IJCAI, 2016.

\bibitem{b12} Z. Liu, P. Luo, X. Wang, and X. Tang, ``Deep learning face attributes in the wild,'' ICCV, 2015.

\bibitem{b13} Y. Zhong, J. Sullivan, and H. Li, ``Leveraging mid-level deep representations for predicting face attributes in the wild,'' ICIP, 2016.

\bibitem{b14} E. M. Hand, and R. Chellappa, ``Attributes for improved attributes: a multi-mask Network for attribute classification,'' arxiv, 2016.

\bibitem{b15} X. Hou, L. Shen, K. Sun, and G. Qiu, ``Deep Feature Consistent Variational Autoencoder,'' WACV, 2017.

\bibitem{b16} R. Ranjan, S. Sankaranarayanan, C. D. Castillo, and Rama Chellappa, ``An all-in-one convolutional neural network for face analysis,'' FG, 2017.

\bibitem{b17} Y. Zhong, J. Sullivan, and H. Li, ``Face attribute prediction using off-the-shelf deep learning networks,'' CoRR, 2016.

\bibitem{b18} R. Torfason, E. Agustsson, R. Rothe, and R. Timofte, ``From face images and attributes to attributes,'' ACCV, 2016.

\bibitem{2b18} Y. Lu, A. Kumar, S. Zhai, Y. Cheng, T.Javidi, R. Feris, ``Fully-adaptive feature sharing in multi-task networks with applications in person attribute classification,'' CVPR, 2016.


\bibitem{b19} D. Gao, P. Yuan, N. Sun, X. Wu, and Y. Cai, ``Face attribute prediction with convolutional neural networks,'' ROBIO, 2017.


\bibitem{2b19} E. M. Hand, C. Castillo and R. Chellappa, ``Doing the Best We Can with What We Have:
Multi-Label Balancing with Selective Learning for Attribute Prediction,'' AAAI, 2018.

\bibitem{3b19} H. Han, A. Jain, F. Wang, S. Shan, X. Chen, ``Heterogeneous Face Attribute Estimation:
A Deep Multi-Task Learning Approach,'' TPAMI, 2018.

\bibitem{b20} S. Escalera, J. Fabian, P. Pardo, X. Baro, J. Gonzalez, H. J. Escalante, D. Misevic, U. Steiner, and I. Guyon, ``ChaLearn Looking at People 2015: Apparent Age and Cultural Event Recognition Datasets and Results,'' ICCV, 2015.

\bibitem{b21} https://www.crummy.com/software/BeautifulSoup/

\bibitem{b22} https://www.seleniumhq.org/projects/webdriver/

\bibitem{b23} https://github.com/ageitgey/face\_recognition/

\bibitem{2b23} https://keras.io/

\bibitem{b24} K. Simonyan, and A. Zisserman, ``Very deep covolutional networks
for large-scale image recognition,'' ICLR, 2015. 

\bibitem{b25} K. He, X. Zhang, S. Ren, and J. Sun, ``Deep residual learning for image recognition,'' CVPR, 2016.

\bibitem{b26} https://github.com/rcmalli/keras-vggface

\bibitem{b27}  X. Glorot and Y. Bengio, ``Understanding the difficulty of training
deep feedforward neural networks,'' AISTATS, 2010.

\bibitem{b28} D. Yi, Z. Lei, S. Liao, and S. Z. Li., ``Learning face representation from scratch,'' CoRR, 2014.

\bibitem{b29} K. Ricanek Jr.,  and T. Tesafaye ``Morph: A longitudinal image database
of normal adult age-progression,'' FGR, 2006.

\bibitem{b30} G. Levi, and T. Hassner ``Age and gender classification using convolutional
neural networks,'' CVPR, 2015.

\bibitem{b31} M. Kostinger, P. Wohlhart, P. Roth, and H. Bischof ``Annotated facial
landmarks in the wild: A large-scale, real-world database for facial
landmark localization,'' ICCV, 2011.

\end{thebibliography}
\end{document}